\documentclass[runningheads]{llncs}
\usepackage{graphicx}
\usepackage{amsmath,amssymb} 
\usepackage{color}
\usepackage[width=122mm,left=12mm,paperwidth=146mm,height=193mm,top=12mm,paperheight=217mm]{geometry}
\usepackage{adjustbox}
\usepackage{multirow}
\usepackage{subfigure}
\usepackage{epsfig,epstopdf}
\usepackage{bbding}
\usepackage{cite}
\usepackage[colorlinks]{hyperref}

\newcommand{\widthscalefive}{0.145}
\begin{document}
\pagestyle{headings}
\mainmatter
\def\ECCV18SubNumber{642}  

\title{Image Super-Resolution Using Very Deep Residual Channel Attention Networks} 

\titlerunning{Image Super-Resolution Using Very Deep RCAN}

\authorrunning{Yulun Zhang \textit{et al.}}

\author{Yulun Zhang$^{1}$, Kunpeng Li$^{1}$, Kai Li$^{1}$, Lichen Wang$^{1}$,\\ Bineng Zhong$^{1}$, and Yun Fu$^{1,2}$}
\institute{$^{1}$Department of ECE, Northeastern University, Boston, USA\\
$^{2}$College of Computer and Information Science, Northeastern University, Boston, USA\\
\email{ \{yulun100,li.kai.gml,wanglichenxj\}@gmail.com,\\ bnzhong@hqu.edu.cn,  \{kunpengli,yunfu\}@ece.neu.edu}}

\maketitle

\begin{abstract}
Convolutional neural network (CNN) depth is of crucial importance for image super-resolution (SR). However, we observe that deeper networks for image SR are more difficult to train. The low-resolution inputs and features contain abundant low-frequency information, which is treated equally across channels, hence hindering the representational ability of CNNs. To solve these problems, we propose the very deep residual channel attention networks (RCAN). Specifically, we propose a residual in residual (RIR) structure to form very deep network, which consists of several residual groups with long skip connections. Each residual group contains some residual blocks with short skip connections. Meanwhile, RIR allows abundant low-frequency information to be bypassed through multiple skip connections, making the main network focus on learning high-frequency information. Furthermore, we propose a channel attention mechanism to adaptively rescale channel-wise features by considering interdependencies among channels. Extensive experiments show that our RCAN achieves better accuracy and visual improvements against state-of-the-art methods.
\keywords{Super-Resolution, Residual in Residual, Channel Attention}
\end{abstract}

\begin{figure}[t]
	\newlength\fsfourteen
	\setlength{\fsfourteen}{-1.3mm}
	\scriptsize
	\centering
	\begin{tabular}{cc}
	\tiny
		\begin{adjustbox}{valign=t}
			\begin{tabular}{ccccccc}
				\includegraphics[width=0.138\textwidth]{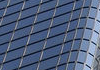} \hspace{\fsfourteen} &
				\includegraphics[width=0.138\textwidth]{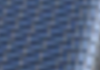}  \hspace{\fsfourteen} &
				\includegraphics[width=0.138\textwidth]{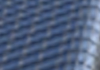} \hspace{\fsfourteen} &
				\includegraphics[width=0.138\textwidth]{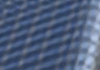} \hspace{\fsfourteen} &
				\includegraphics[width=0.138\textwidth]{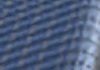} \hspace{\fsfourteen} &
				\includegraphics[width=0.138\textwidth]{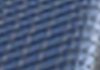}
				\hspace{\fsfourteen} &
				\includegraphics[width=0.138\textwidth]{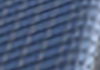}
				\\
				HR \hspace{\fsfourteen} &
				Bicubic \hspace{\fsfourteen} &
				SRCNN~\cite{dong2016image} \hspace{\fsfourteen} &
				FSRCNN~\cite{dong2016accelerating} \hspace{\fsfourteen} &
				SCN~\cite{wang2015deep} \hspace{\fsfourteen} &
				VDSR~\cite{kim2016accurate} &
				DRRN~\cite{tai2017image} \hspace{\fsfourteen} 
				\\
				\includegraphics[width=0.138\textwidth]{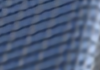} \hspace{\fsfourteen} &
				\includegraphics[width=0.138\textwidth]{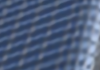} \hspace{\fsfourteen} &
				\includegraphics[width=0.138\textwidth]{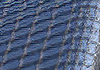} \hspace{\fsfourteen} &
				\includegraphics[width=0.138\textwidth]{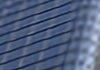} \hspace{\fsfourteen} &
				\includegraphics[width=0.138\textwidth]{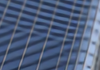} \hspace{\fsfourteen} &
				\includegraphics[width=0.138\textwidth]{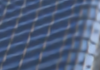} \hspace{\fsfourteen} &
				\includegraphics[width=0.138\textwidth]{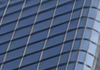} 
				\\ 
				LapSRN~\cite{lai2017deep} \hspace{\fsfourteen} &
				MSLapSRN~\cite{MSLapSRN} \hspace{\fsfourteen} &
				ENet-PAT~\cite{sajjadi2017enhancenet} \hspace{\fsfourteen} &
				MemNet~\cite{tai2017memnet} \hspace{\fsfourteen} &
				EDSR~\cite{lim2017enhanced} \hspace{\fsfourteen} &
				SRMDNF~\cite{zhang2018learning} \hspace{\fsfourteen} &
				RCAN (ours)
				\\
			\end{tabular}
		\end{adjustbox}

	\end{tabular}
	\vspace{-3mm}
	\caption{
		Visual results with Bicubic (BI) degradation (4$\times$) on ``img\_074" from Urban100
	}
	\label{fig:result_4x_14}
\vspace{-6mm}
\end{figure}
\vspace{-8mm}
\section{Introduction}
\label{sec:introduction}
\vspace{-2mm}
We address the problem of reconstructing an accurate high-resolution (HR) image given its low-resolution (LR) counterpart, usually referred as single image super-resolution (SR)~\cite{freeman2000learning}. Image SR is used in various computer vision applications, ranging from security and surveillance imaging~\cite{zou2012very}, medical imaging~\cite{shi2013cardiac} to object recognition~\cite{sajjadi2017enhancenet}. However, image SR is an ill-posed problem, since there exists 
multiple solutions for any LR input. To tackle such an inverse problem, numerous learning based methods have been proposed to learn mappings between LR and HR image pairs.

Recently, deep convolutional neural network (CNN) based methods~\cite{dong2016image,dong2016accelerating,wang2015deep,kim2016accurate,tai2017image,lai2017deep,MSLapSRN,sajjadi2017enhancenet,tai2017memnet,lim2017enhanced,zhang2017learning,zhang2018learning,haris2018deep,zhang2018residual} have achieved significant improvements over conventional SR methods. Among them, Dong et al.~\cite{dong2014learning} proposed SRCNN by firstly introducing a three-layer CNN for image SR. Kim et al. increased the network depth to 20 in VDSR~\cite{kim2016accurate} and DRCN~\cite{kim2016deeply}, achieving notable improvements over SRCNN. Network depth was demonstrated to be of central importance for many visual recognition tasks, especially when He at al.~\cite{he2016deep} proposed residual net (ResNet), which reaches 1,000 layers with residual blocks. Such effective residual learning strategy was then introduced in many other CNN-based image SR methods~\cite{tai2017image,tai2017memnet,ledig2017photo,lim2017enhanced,sajjadi2017enhancenet}. Lim et al.~\cite{lim2017enhanced} built a very wide network EDSR and a very deep one MDSR (about 165 layers) by using simplified residual blocks. The great improvements on performance of EDSR and MDSR indicate that the depth of representation is of crucial importance for image SR. However, to the best of our knowledge, simply stacking residual blocks to construct deeper networks can hardly obtain better improvements. Whether deeper networks can further contribute to image SR and how to construct very deep trainable networks remains to be explored. 

On the other hand, most recent CNN-based methods~\cite{dong2016image,dong2016accelerating,wang2015deep,kim2016accurate,tai2017image,lai2017deep,MSLapSRN,sajjadi2017enhancenet,tai2017memnet,lim2017enhanced,zhang2018learning} treat channel-wise features equally, which lacks flexibility in dealing with different types of information (e.g., low- and high-frequency information). Image SR can be viewed as a process, where we try to recover as more high-frequency information as possible. The LR images contain most low-frequency information, which can directly forwarded to the final HR outputs and don't need too much computation. While, the leading CNN-based methods (e.g., EDSR~\cite{lim2017enhanced}) would extract features from the original LR inputs and treat each channel-wise feature equally. Such process would wastes unnecessary computations for abundant low-frequency features, lacks discriminative learning ability across feature channels, and finally hinders the representational power of deep networks.    


To practically resolve these problems, we propose a residual channel attention network (RCAN) to obtain very deep trainable network and adaptively learn more useful channel-wise features simultaneously. To ease the training of very deep networks (e.g., over 400 layers), we propose residual in residual (RIR) structure, where the residual group (RG) serves as the basic module and long skip connection (LSC) allows residual learning in a coarse level. In each RG module, we stack several simplified residual block~\cite{lim2017enhanced} with short skip connection (SSC). The long and short skip connection as well as the short-cut in residual block allow abundant low-frequency information to be bypassed through these identity-based skip connections, which can ease the flow of information. To make a further step, we propose channel attention (CA) mechanism to adaptively rescale each channel-wise feature by modeling the interdependencies across feature channels. Such CA mechanism allows our proposed network to concentrate on more useful channels and enhance discriminative learning ability. As shown in Figure~\ref{fig:result_4x_14}, our RCAN achieves better visual SR result compared with state-of-the-art methods.

Overall, our contributions are three-fold: (1) We propose the very deep residual channel attention networks (RCAN) for highly accurate image SR. Our RCAN can reach much deeper than previous CNN-based methods and obtains much better SR performance. (2) We propose residual in residual (RIR) structure to construct very deep trainable networks. The long and short skip connections in RIR help to bypass abundant low-frequency information and make the main network learn more effective information. (3) We propose channel attention (CA) mechanism to adaptively rescale features by considering interdependencies among feature channels. Such CA mechanism further improves the representational ability of the network.
\vspace{-4mm}
\section{Related Work}
\vspace{-3mm}
Numerous image SR methods have been studied in the computer vision community~\cite{dong2016image,dong2016accelerating,wang2015deep,kim2016accurate,tai2017image,lai2017deep,MSLapSRN,sajjadi2017enhancenet,tai2017memnet,lim2017enhanced,zhang2018learning,huang2015single}. Attention mechanism is popular in high-level vision tasks, but is seldom investigated in low-level vision applications~\cite{hu2017squeeze}. Due to space limitation, here we focus on works related to CNN-based methods and attention mechanism. 

\textbf{Deep CNN for SR.} The pioneer work was done by Dong et al.~\cite{dong2014learning}, who proposed SRCNN for image SR and achieved superior performance against previous works. By introducing residual learning to ease the training difficulty, Kim et al. proposed VDSR~\cite{kim2016accurate} and DRCN~\cite{kim2016deeply} with 20 layers and achieved significant improvement in accuracy. Tai et al. later introduced recursive blocks in DRRN~\cite{tai2017image} and memory block in MemNet~\cite{tai2017memnet}. These methods would have to first interpolate the LR inputs to the desired size, which inevitably loses some details and increases computation greatly. 

Extracting features from the original LR inputs and upscaling spatial resolution at the network tail then became the main choice for deep architecture. A faster network structure FSRCNN~\cite{dong2016accelerating} was proposed to accelerate the training and testing of SRCNN. Ledig et al.~\cite{ledig2017photo} introduced ResNet~\cite{he2016deep} to construct a deeper network, SRResNet, for image SR. They also proposed SRGAN with perceptual losses~\cite{johnson2016perceptual} and generative adversarial network (GAN)~\cite{goodfellow2014generative} for photo-realistic SR. Such GAN based model was then introduced in EnhanceNet~\cite{sajjadi2017enhancenet}, which combines automated texture synthesis and perceptual loss. Although SRGAN and Enhancenet can alleviate the blurring and oversmoothing artifacts to some degree, their predicted results may not be faithfully reconstructed and produce unpleasing artifacts. By removing unnecessary modules in conventional residual networks, Lim et al.~\cite{lim2017enhanced} proposed EDSR and MDSR, which achieve significant improvement. However, most of these methods have limited network depth, which has demonstrated to be very important in visual recognition tasks~\cite{he2016deep} and can reach to about 1,000 layers. Simply stacking residual blocks in MDSR~\cite{lim2017enhanced}, very deep networks can hardly achieved improvements. Furthermore, most of these methods treat the channel-wise features equally, hindering better discriminative ability for different types of features. 


\textbf{Attention mechanism.} Generally, attention can be viewed as a guidance to bias the allocation of available processing resources towards the most informative components of an input~\cite{hu2017squeeze}. Recently, tentative works have been proposed to apply attention into deep neural networks~\cite{wang2017residual,hu2017squeeze,li2018tell}, ranging from localization and understanding in images~\cite{cao2015look,jaderberg2015spatial} to sequence-based networks~\cite{bluche2016joint,miech2017learnable}. It's usually combined with a gating function (e.g., sigmoid) to rescale the feature maps. Wang et al.~\cite{wang2017residual} proposed residual attention network for image classification with a trunk-and-mask attention mechanism. Hu et al.~\cite{hu2017squeeze} proposed squeeze-and-excitation (SE) block to model channel-wise relationships to obtain significant performance improvement for image classification. However, few works have been proposed to investigate the effect of attention for low-level vision tasks (e.g., image SR). 

In image SR, high-frequency channel-wise features are more informative for HR reconstruction. If our network pays more attention to such channel-wise features, it should be promising to obtain improvements. To investigate such mechanism in very deep CNN, we propose very deep residual channel attention networks (RCAN), which we will detail in next section.

\begin{figure}[t]
\centering
\includegraphics[scale=0.62]{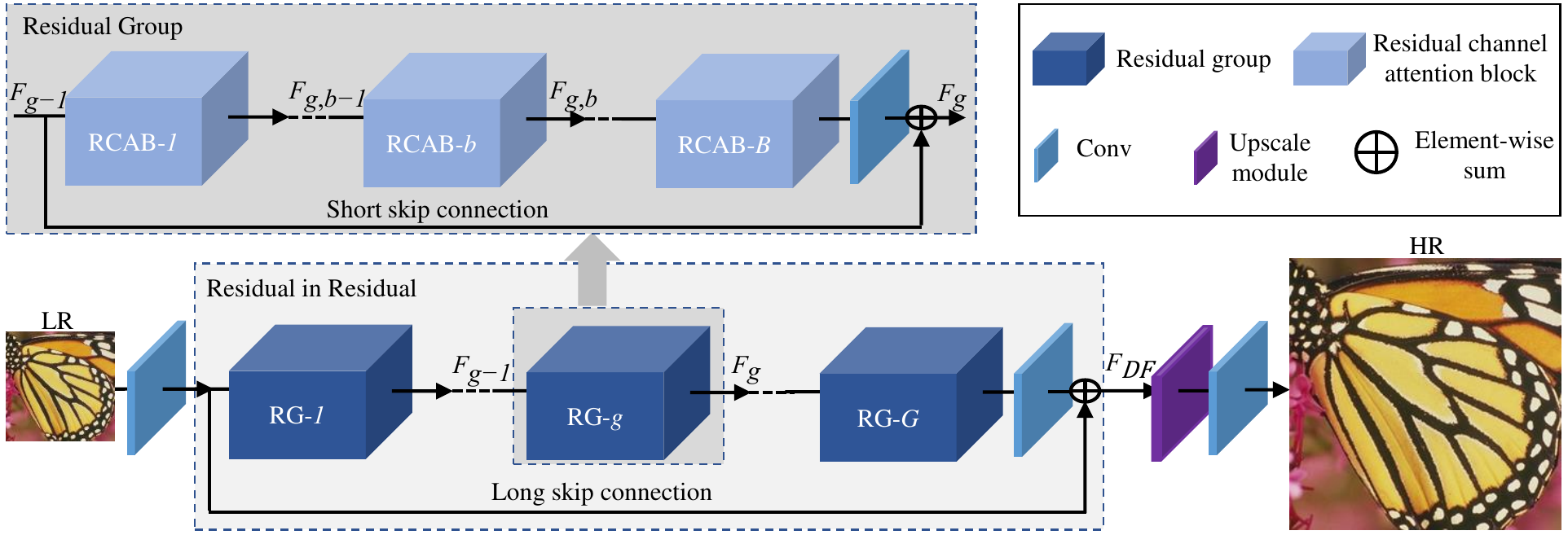}
\vspace{-8mm}
\caption{Network architecture of our residual channel attention network (RCAN)}
\label{fig:network_RCA}
\vspace{-4mm}
\end{figure}

\vspace{-4mm}
\section{Residual Channel Attention Network (RCAN)}
\vspace{-2mm}
\subsection{Network Architecture}
\vspace{-2mm}
As shown in Figure~\ref{fig:network_RCA}, our RCAN mainly consists four parts: shallow feature extraction, residual in residual (RIR) deep feature extraction, upscale module, and reconstruction part. Let's denote $I_{LR}$ and $I_{SR}$ as the input and output of RCAN. As investigated in~\cite{ledig2017photo,lim2017enhanced}, we use only one convolutional layer (Conv) to extract the shallow feature $F_{0}$ from the LR input
\begin{align}
\begin{split}
F_{0}=H_{SF}\left ( I_{LR} \right ),
\end{split}
\end{align}
where $H_{SF}\left ( \cdot  \right )$ denotes convolution operation. $F_{0}$ is then used for deep feature extraction with RIR module. So we can further have
\begin{align}
\begin{split}
F_{DF}=H_{RIR}\left ( F_{0} \right ),
\end{split}
\end{align}
where $H_{RIR}\left ( \cdot  \right )$ denotes our proposed very deep residual in residual structure, which contains $G$ residual groups (RG). To the best of our knowledge, our proposed RIR achieves the largest depth so far and provides very large receptive field size. So we treat its output as deep feature, which is then upscaled via a upscale module
\begin{align}
\begin{split}
F_{UP}=H_{UP}\left ( F_{DF} \right ),
\end{split}
\end{align}
where $H_{UP}\left ( \cdot  \right )$ and $F_{UP}$ denote a upscale module and upscaled feature respectively. 

There're several choices to serve as upscale modules, such as deconvolution layer (also known as transposed convolution)~\cite{dong2016accelerating}, nearest-neighbor upsampling + convolution~\cite{dumoulin2016learned}, and ESPCN~\cite{shi2016real}. Such post-upscaling strategy has been demonstrated to be more efficient for both computation complexity and achieve higher performance than pre-upscaling SR methods (e.g., DRRN~\cite{tai2017image} and MemNet~\cite{tai2017memnet}). The upscaled feature is then reconstructed via one Conv layer
\begin{align}
\begin{split}
I_{SR}=H_{REC}\left ( F_{UP} \right )=H_{RCAN}\left ( I_{LR} \right ),
\end{split}
\end{align}
where $H_{REC}\left ( \cdot  \right )$ and $H_{RCAN}\left ( \cdot  \right )$ denote the reconstruction layer and the function of our RCAN respectively. 

Then RCAN is optimized with loss function. Several loss functions have been investigated, such as $L_{2}$~\cite{dong2016image,dong2016accelerating,wang2015deep,kim2016accurate,tai2017image,sajjadi2017enhancenet,tai2017memnet,zhang2018learning,haris2018deep}, $L_{1}$~\cite{lim2017enhanced,lai2017deep,MSLapSRN,zhang2018residual}, perceptual and adversarial losses~\cite{sajjadi2017enhancenet,ledig2017photo}. To show the effectiveness of our RCAN, we choose to optimize same loss function as previous works (e.g., $L_{1}$ loss function). Given a training set $\left \{ I_{LR}^{i}, I_{HR}^{i}\right \}_{i=1}^{N}$, which contains $N$ LR inputs and their HR counterparts. The goal of training RCAN is to minimize the $L_1$ loss function
\begin{align}
\begin{split}
L\left ( \Theta  \right )=\frac{1}{N}\sum_{i=1}^{N}\left \| H_{RCAN}\left ( I_{LR}^{ i } \right )-I_{HR}^{ i } \right \|_{1},
\end{split}
\end{align}
where $\Theta$ denotes the parameter set of our network. The loss function is optimized by using stochastic gradient descent. More details of training would be shown in Section~\ref{subsec:settings}. As we choose the shallow feature extraction $H_{SF}\left ( \cdot  \right )$, upscaling module $H_{UP}\left ( \cdot  \right )$, and reconstruction part $H_{UP}\left ( \cdot  \right )$ as similar as previous works (e.g., EDSR~\cite{lim2017enhanced} and RDN~\cite{zhang2018residual}), we pay more attention to our proposed RIR, CA, and the basic module RCAB.  

\subsection{Residual in Residual (RIR)}
We now give more details about our proposed RIR structure (see Figure~\ref{fig:network_RCA}), which contains $G$ residual groups (RG) and long skip connection (LSC). Each RG further contains $B$ residual channel attention blocks (RCAB) with short skip connection (SSC). Such residual in residual structure allows to train very deep CNN (over 400 layers) for image SR with high performance.

It has been demonstrated that stacked residual blocks and LSC can be used to construct deep CNN in~\cite{lim2017enhanced}. In visual recognition, residual blocks~\cite{he2016deep} can be stacked to achieve more than 1,000-layer trainable networks. However, in image SR, very deep network built in such way would suffer from training difficulty and can hardly achieve more performance gain. Inspired by previous works in SRRestNet~\cite{ledig2017photo} and EDSR~\cite{lim2017enhanced}, we proposed residual group (RG) as the basic module for deeper networks. A RG in the $g$-th group is formulated as
\begin{align}
\begin{split}
F_{g} = H_{g}\left ( F_{g-1} \right )=H_{g}\left ( H_{g-1}\left ( \cdots H_{1}\left ( F_{0} \right ) \cdots  \right ) \right ),
\end{split}
\end{align} 
where $H_{g}$ denotes the function of $g$-th RG. $F_{g-1}$ and $F_{g}$ are the input and output for $g$-th RG. We observe that simply stacking many RGs would fail to achieve better performance. To solve the problem, the long skip connection (LSC) is further introduced in RIR to stabilize the training of very deep network. LSC also makes better performance possible with residual learning via
\begin{align}
\begin{split}
F_{DF}=F_{0}+W_{LSC}F_{G}=F_{0}+W_{LSC}H_{g}\left ( H_{g-1}\left ( \cdots H_{1}\left ( F_{0} \right ) \cdots  \right ) \right ),
\end{split}
\end{align} 
where $W_{LSC}$ is the weight set to the Conv layer at the tail of RIR. The bias term is omitted for simplicity. LSC can not only ease the flow of information across RGs, but only make it possible for RIR to learning residual information in a coarse level. 

As discussed in Section~\ref{sec:introduction}, there are lots of abundant information in the LR inputs and features and the goal of SR network is to recover more useful information. The abundant low-frequency information can be bypassed through identity-based skip connection. To make a further step towards residual learning, we stack $B$ residual channel attention blocks in each RG. The $b$-th residual channel attention block (RCAB) in $g$-th RG can be formulated as
\begin{align}
\begin{split}
F_{g,b}=H_{g,b}\left ( F_{g,b-1} \right )=H_{g,b}\left ( H_{g,b-1}\left ( \cdots H_{g,1}\left ( F_{g-1} \right ) \cdots  \right ) \right ),
\end{split}
\end{align} 
where $F_{g,b-1}$ and $F_{g,b}$ are the input and output of the $b$-th RCAB in $g$-th RG. The corresponding function is denoted with $H_{g,b}$. To make the main network pay more attention to more informative features, a short skip connection (SSC) is introduced to obtain the block output via
\begin{align}
\begin{split}
F_{g}=F_{g-1}+W_{g}F_{g,B}=F_{g-1}+W_{g}H_{g,B}\left ( H_{g,B-1}\left ( \cdots H_{g,1}\left ( F_{g-1} \right ) \cdots  \right ) \right ),
\end{split}
\end{align} 
where $W_{g}$ is the weight set to the Conv layer at the tail of $g$-th RG. The SSC further allows the main parts of network to learn residual information. With LSC and SSC, more abundant low-frequency information is easier bypassed in the training process. To make a further step towards more discriminative learning, we pay more attention to channel-wise feature rescaling with channel attention. 
\begin{figure}[t]
\centerline{
\includegraphics[scale = 1]{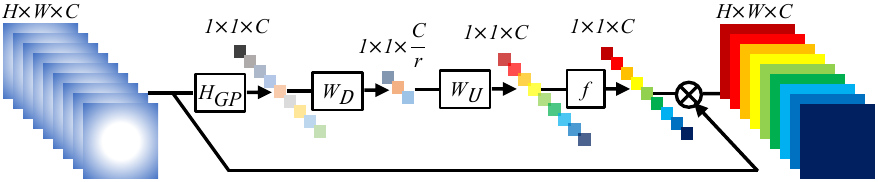}
}
\vspace{-3mm}
\caption{Channel attention (CA). $\otimes$ denotes element-wise product}
\label{fig:CA}
\vspace{-5mm}
\end{figure}
\vspace{-6mm}
\subsection{Channel Attention (CA)}
Previous CNN-based SR methods treat LR channel-wise features equally, which is not flexible for the real cases. In order to make the network focus on more informative features, we exploit the interdependencies among feature channels, resulting in a channel attention (CA) mechanism (see Figure~\ref{fig:CA}). 

How to generate different attention for each channel-wise feature is a key step. Here we mainly have two concerns: First, information in the LR space has abundant low-frequency and valuable high-frequency components. The low-frequency parts seem to be more complanate. The high-frequency components would usually be regions, being full of edges, texture, and other details. On the other hand, each filter in Conv layer operates with a local receptive field. Consequently, the output after convolution is unable to exploit contextual information outside of the local region. 

Based on these analyses, we take the channel-wise global spatial information into a channel descriptor by using global average pooling. As shown in Figure~\ref{fig:CA}, let $X=\left [ x_1,\cdots , x_c,\cdots ,x_C \right ]$ be an input, which has $C$ feature maps with size of $H\times W$. The channel-wise statistic $z\in \mathbb{R}^{C}$ can be obtained by shrinking $X$ through spatial dimensions $H\times W$. Then the $c$-th element of $z$ is determined by
\begin{align}
\begin{split}
z_{c}=H_{GP}\left ( x_{c} \right )=\frac{1}{H\times W}\sum_{i=1}^{H}\sum_{j=1}^{W}x_{c}\left ( i,j \right ),
\end{split}
\end{align} 
where $x_{c}\left ( i,j \right )$ is the value at position $\left ( i,j \right )$ of $c$-th feature $x_c$. $H_{GP}\left ( \cdot \right )$ denotes the global pooling function. Such channel statistic can be viewed as a collection of the local descriptors, whose statistics contribute to express the whole image~\cite{hu2017squeeze}. Except for global average pooling, more sophisticated aggregation techniques could also be introduced here.

To fully capture channel-wise dependencies from the aggregated information by global average pooling, we introduce a gating mechanism. As discussed in~\cite{hu2017squeeze}, the gating mechanism should meet two criteria: First, it must be able to learn nonlinear interactions between channels. Second, as multiple channel-wise features can be emphasized opposed to one-hot activation, it must learn a non-mututually-exclusive relationship. Here, we opt to exploit simple gating mechanism with sigmoid function
\begin{align}
\begin{split}
s = f \left ( W_{U}\delta \left ( W_{D}z \right ) \right ),
\end{split}
\end{align}
where $f\left ( \cdot \right )$ and $\delta \left ( \cdot \right )$ denote the sigmoid gating and ReLU~\cite{nair2010rectified} function, respectively. $W_{D}$ is the weight set of a Conv layer, which acts as channel-downscaling with reduction ratio $r$. After being activated by ReLU, the low-dimension signal is then increased with ratio $r$ by a channel-upscaling layer, whose weight set is $W_{U}$. Then we obtain the final channel statistics $s$, which is used to rescale the input $x_{c}$ 
\begin{align}
\begin{split}
\widehat{x}_c=s_c\cdot x_{c},
\end{split}
\end{align}  
where $s_{c}$ and $x_{c}$ are the scaling factor and feature map in the $c$-th channel. With channel attention, the residual component in the RCAB is adaptively rescaled.
\begin{figure}[t]
\centerline{
\includegraphics[scale = 0.6]{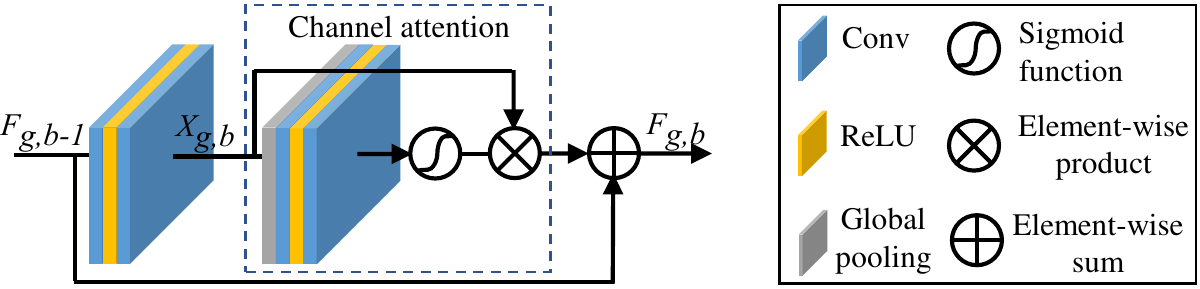}}
\vspace{-3mm}
\caption{Residual channel attention block (RCAB)}

\label{fig:RCAB}
\vspace{-6mm}
\end{figure}
\vspace{-5mm}
\subsection{Residual Channel Attention Block (RCAB)}
\vspace{-2mm}
As discussed above, residual groups and long skip connection allow the main parts of network to focus on more informative components of the LR features. Channel attention extracts the channel statistic among channels to further enhance the discriminative ability of the network.

At the same time, inspired by the success of residual blocks (RB) in~\cite{lim2017enhanced}, we integrate CA into RB and propose residual channel attention block (RCAB) ( see Figure~\ref{fig:RCAB}). For the $b$-th RB in $g$-th RG, we have
\begin{align}
\begin{split}
F_{g,b}=F_{g,b-1}+R_{g,b}\left ( X_{g,b} \right )\cdot X_{g,{b}},
\end{split}
\end{align} 
where $R_{g,b}$ denotes the function of channel attention. $F_{g,b}$ and $F_{g,b-1}$ are the input and output of RCAB, which learns the residual $X_{g,b}$ from the input. The residual component is mainly obtained by two stacked Conv layers 
\begin{align}
\begin{split}
X_{g,b}=W_{g,b}^{2}\delta \left ( W_{g,b}^{1}F_{g,b-1} \right ),
\end{split}
\end{align} 
where $W_{g,b}^{1}$ and $W_{g,b}^{2}$ are weight sets the two stacked Conv layers in RCAB.

We further show the relationships between our proposed RCAB and residual block (RB) in~\cite{lim2017enhanced}. We find that the RBs used in MDSR and EDSR~\cite{lim2017enhanced} can be viewed as special cases of our RCAB. For RB in MDSR, there is no rescaling operation. It is the same as RCAB, where we set $R_{g,b}\left ( \cdot \right )$ as constant 1. For RB with constant rescaling (e.g., 0.1) in EDSR, it is the same as RCAB with $R_{g,b}\left ( \cdot \right )$ set to be 0.1. Although the channel-wise feature rescaling is introduced to train a very wide network, the interdependencies among channels are not considered in EDSR. In these cases, the CA is not considered.

Based on residual channel attention block (RCAB) and RIR structure, we construct a very deep RCAN for highly accurate image SR and achieve notable performance improvements over previous leading methods. More discussions about the effects of each proposed component are shown in Section~\ref{subsec:ablation_study}. 

\vspace{-4mm}
\subsection{Implementation Details}
\vspace{-2mm}
Now we specify the implementation details of our proposed RCAN. We set RG number as $G$=10 in the RIR structure. In each RG, we set RCAB number as 20. We set 3$\times$3 as the size of all Conv layers except for that in the channel-downscaling and channel-upscaling, whose kernel size is 1$\times$1. For Conv layers with kernel size 3$\times$3, zero-padding strategy is used to keep size fixed. Conv layers in shallow feature extraction and RIR structure have $C$=64 filters, except for that in the channel-downscaling. Conv layer in channel-downscaling has $\frac{C}{r}$=4 filters, where the reduction ratio $r$ is set as 16. For upscaling module $H_{UP}\left ( \cdot  \right )$, we follow~\cite{shi2016real,lim2017enhanced,zhang2018residual} and use ESPCNN~\cite{shi2016real} to upscale the coarse resolution features to fine ones. The final Conv layer has 3 filters, as we output color images. While, our network can also process gray images.

\begin{table}[tbp]
\centering
\begin{center}
\caption{Investigations of RIR (including LSC and SSC) and CA. We observe the best PSNR (dB) values on Set5 (2$\times$) in 5$\times$10$^{4}$ iterations}
\label{tab:results_ablation} 
\vspace{-3mm}
\begin{tabular}{|c|c|c|c|c|c|c|c|c|c|c|c|}

\hline
\multirow{2}{*}{Residual in Residual (RIR)} & LSC  & \XSolidBrush & \Checkmark & \XSolidBrush & \Checkmark & \XSolidBrush & \Checkmark & \XSolidBrush & \Checkmark 
\\
\cline{2-2}
& SSC & \XSolidBrush & \XSolidBrush   & \Checkmark & \Checkmark & \XSolidBrush & \XSolidBrush   & \Checkmark & \Checkmark
\\
\cline{1-2}
\multicolumn{2}{|c|}{Channel attention (CA)} & \XSolidBrush & \XSolidBrush   & \XSolidBrush & \XSolidBrush & \Checkmark & \Checkmark & \Checkmark & \Checkmark 
\\
\hline
\hline
\multicolumn{2}{|c|}{PSNR on Set5 (2$\times$)} & 37.45  & 37.77  & 37.81 & 37.87 & 37.52 & 37.85 & 37.86 & 37.90
\\
\hline
\end{tabular}
\end{center}
\vspace{-10mm}
\end{table}

\section{Experiments}
\vspace{-3mm}
\subsection{Settings}
\label{subsec:settings}
\vspace{-2mm}
We clarify the experimental settings about datasets, degradation models, evaluation metric, and training settings. 

\textbf{Datasets and degradation models.} Following~\cite{timofte2017ntire,lim2017enhanced,zhang2018residual,zhang2018learning}, we use 800 training images from DIV2K dataset~\cite{timofte2017ntire} as training set. For testing, we use five standard benchmark datasets: Set5~\cite{bevilacqua2012low}, Set14~\cite{zeyde2012single}, B100~\cite{martin2001database}, Urban100~\cite{huang2015single}, and Manga109~\cite{matsui2017sketch}. We conduct experiments with Bicubic (BI) and blur-downscale (BD) degradation models~\cite{zhang2017learning,zhang2018learning,zhang2018residual}.

\textbf{Evaluation metrics.} The SR results are evaluated with PSNR and SSIM~\cite{wang2004image} on Y channel (i.e., luminance) of transformed YCbCr space. We also provide performance (e.g., top-1 and top-5 recognition errors) comparisons on object recognition by several leading SR methods. 

\textbf{Training settings.} Data augmentation is performed on the 800 training images, which are randomly rotated by 90$^{\circ}$, 180$^{\circ}$, 270$^{\circ}$ and flipped horizontally. In each training batch, 16 LR color patches with the size of $48\times48$ are extracted as inputs. Our model is trained by ADAM optimizor~\cite{kingma2014adam} with $\beta_{1}=0.9$, $\beta_{2}=0.999$, and $\epsilon=10^{-8}$. The initial leaning rate is set to $10^{-4}$ and then decreases to half every $2\times10^{5}$ iterations of back-propagation. We use PyTorch~\cite{paszke2017automatic} to implement our models with a Titan Xp GPU.\footnote{The RCAN source code is available at \href{https://github.com/yulunzhang/RCAN}{https://github.com/yulunzhang/RCAN}.}

\vspace{-4mm}
\subsection{Effects of RIR and CA}
\label{subsec:ablation_study}
\vspace{-2mm}
We study the effects of residual in residual (RIR) and channel attention (CA).

\textbf{Residual in residual (RIR).} To demonstrate the effect of our proposed residual in residual structure, we remove long skip connection (LSC) or/and short skip connection (SSC) from very deep networks. Specifically, we set the number of residual block as 200, namely 10 residual groups, resulting in very deep networks with over 400 Conv layers. In Table~\ref{tab:results_ablation}, when both LSC and SSC are removed, the PSNR value on Set5 ($\times2$) is relatively low, no matter channel attention (CA) is used or not. For example, in the first column, the PSNR is 37.45 dB. After adding RIR, the performance reaches 37.87 dB. When CA is added, the performance can be improved from 37.52 dB to 37.90 dB by using RIR. This indicates that simply stacking residual blocks is not applicable to achieve very deep and powerful networks for image SR. The performance would increase with LSC or SSC and can obtain better results by using both of them. These comparisons show that LSC and SSC are essential for very deep networks. They also demonstrate the effectiveness of our proposed residual in residual (RIR) structure for very deep networks. 

\textbf{Channel attention (CA).} We further show the effect of channel attention (CA) based on the observations and discussions above. When we compare the results of first 4 columns and last 4 columns, we find that networks with CA would perform better than those without CA. Benefitting from very large network depth, the very deep trainable networks can achieve a very high performance. It's hard to obtain further improvements from such deep networks, but we obtain improvements with CA. Even without RIR, CA can improve the performance from 37.45 dB to 37.52 dB. These comparisons firmly demonstrate the effectiveness of CA and indicate adaptive attentions to channel-wise features really improves the performance.     

\vspace{-5mm}
\begin{table}[thbp]
\scriptsize
\center
\begin{center}
\caption{Quantitative results with BI degradation model. Best and second best results are \textbf{highlighted} and \underline{underlined}}
\label{tab:results_psnr_ssim_x2348}
\vspace{-3mm}
\begin{tabular}{|l|c|c|c|c|c|c|c|c|c|c|c|}
\hline
\multirow{2}{*}{Method} & \multirow{2}{*}{Scale} &  \multicolumn{2}{c|}{Set5} &  \multicolumn{2}{c|}{Set14} &  \multicolumn{2}{c|}{B100} &  \multicolumn{2}{c|}{Urban100} &  \multicolumn{2}{c|}{Manga109}  
\\
\cline{3-12}
&  & PSNR & SSIM & PSNR & SSIM & PSNR & SSIM & PSNR & SSIM & PSNR & SSIM 
\\
\hline
\hline
Bicubic & $\times$2 
& 33.66
 & 0.9299
  & 30.24
   & 0.8688
    & 29.56
     & 0.8431
      & 26.88
       & 0.8403
        & 30.80
         & 0.9339
                  
\\
SRCNN~\cite{dong2016image} & $\times$2 
& 36.66
 & 0.9542
  & 32.45
   & 0.9067
    & 31.36
     & 0.8879
      & 29.50
       & 0.8946
        & 35.60
         & 0.9663
                   
\\
FSRCNN~\cite{dong2016accelerating} & $\times$2 
& 37.05
 & 0.9560
  & 32.66
   & 0.9090
    & 31.53
     & 0.8920
      & 29.88
       & 0.9020
        & 36.67
         & 0.9710
                   
\\
VDSR~\cite{kim2016accurate} & $\times$2 
& 37.53
 & 0.9590
  & 33.05
   & 0.9130
    & 31.90
     & 0.8960
      & 30.77
       & 0.9140
        & 37.22
         & 0.9750
                   
\\
LapSRN~\cite{lai2017deep} & $\times$2 
& 37.52
 & 0.9591
  & 33.08
   & 0.9130
    & 31.08
     & 0.8950
      & 30.41
       & 0.9101
        & 37.27
         & 0.9740
                   
\\
MemNet~\cite{tai2017memnet} & $\times$2 
& 37.78
 & 0.9597
  & 33.28
   & 0.9142
    & 32.08
     & 0.8978
      & 31.31
       & 0.9195
        & 37.72
         & 0.9740
                   
\\
EDSR~\cite{lim2017enhanced} & $\times$2 
& 38.11
 & 0.9602
  & 33.92
   & 0.9195
    & 32.32
     & 0.9013
      & 32.93
       & 0.9351
        & 39.10
         & 0.9773
                   
\\
SRMDNF~\cite{zhang2018learning} & $\times$2 
& 37.79
 & 0.9601
  & 33.32
   & 0.9159
    & 32.05
     & 0.8985
      & 31.33
       & 0.9204
        & 38.07
         & 0.9761
                   
\\
D-DBPN~\cite{haris2018deep} & $\times$2 
& 38.09
 & 0.9600
  & 33.85
   & 0.9190
    & 32.27
     & 0.9000
      & 32.55
       & 0.9324
        & 38.89
         & 0.9775        
\\
RDN~\cite{zhang2018residual} & $\times$2 
& 38.24
 & 0.9614
  & 34.01
   & 0.9212
    & 32.34
     & 0.9017
      & 32.89
       & 0.9353
        & 39.18
         & 0.9780
         
\\
RCAN (ours) & $\times$2 
& \underline{38.27}
 & \underline{0.9614}
  & \underline{34.12}
   & \underline{0.9216}
    & \underline{32.41}
     & \underline{0.9027}
      & \underline{33.34}
       & \underline{0.9384}
        & \underline{39.44}
         & \underline{0.9786}

\\
RCAN+ (ours) & $\times$2 
& \textbf{38.33}
 & \textbf{0.9617}
  & \textbf{34.23}
   & \textbf{0.9225}
    & \textbf{32.46}
     & \textbf{0.9031}
      & \textbf{33.54}
       & \textbf{0.9399}
        & \textbf{39.61}
         & \textbf{0.9788}

\\
\hline
\hline
Bicubic & $\times$3 
& 30.39
 & 0.8682
  & 27.55
   & 0.7742
    & 27.21
     & 0.7385
      & 24.46
       & 0.7349
        & 26.95
         & 0.8556
                  
\\
SRCNN~\cite{dong2016image} & $\times$3
& 32.75
 & 0.9090
  & 29.30
   & 0.8215
    & 28.41
     & 0.7863
      & 26.24
       & 0.7989
        & 30.48
         & 0.9117
                    
\\
FSRCNN~\cite{dong2016accelerating} & $\times$3 
& 33.18
 & 0.9140
  & 29.37
   & 0.8240
    & 28.53
     & 0.7910
      & 26.43
       & 0.8080
        & 31.10
         & 0.9210
                   
\\
VDSR~\cite{kim2016accurate} & $\times$3 
& 33.67
 & 0.9210
  & 29.78
   & 0.8320
    & 28.83
     & 0.7990
      & 27.14
       & 0.8290
        & 32.01
         & 0.9340
                   
\\
LapSRN~\cite{lai2017deep} & $\times$3 
& 33.82
 & 0.9227
  & 29.87
   & 0.8320
    & 28.82
     & 0.7980
      & 27.07
       & 0.8280
        & 32.21
         & 0.9350
                   
\\
MemNet~\cite{tai2017memnet} & $\times$3 
& 34.09
 & 0.9248
  & 30.00
   & 0.8350
    & 28.96
     & 0.8001
      & 27.56
       & 0.8376
        & 32.51
         & 0.9369
                   
\\
EDSR~\cite{lim2017enhanced} & $\times$3 
& 34.65
 & 0.9280
  & 30.52
   & 0.8462
    & 29.25
     & 0.8093
      & 28.80
       & 0.8653
        & 34.17
         & 0.9476
                   
\\
SRMDNF~\cite{zhang2018learning} & $\times$3 
& 34.12
 & 0.9254
  & 30.04
   & 0.8382
    & 28.97
     & 0.8025
      & 27.57
       & 0.8398
        & 33.00
         & 0.9403
                   
\\
RDN~\cite{zhang2018residual} & $\times$3 
& 34.71
 & 0.9296
  & 30.57
   & 0.8468
    & 29.26
     & 0.8093
      & 28.80
       & 0.8653
        & 34.13
         & 0.9484
         
\\
RCAN (ours) & $\times$3 
& \underline{34.74}
 & \underline{0.9299}
  & \underline{30.65}
   & \underline{0.8482}
    & \underline{29.32}
     & \underline{0.8111}
      & \underline{29.09}
       & \underline{0.8702}
        & \underline{34.44}
         & \underline{0.9499}

\\
RCAN+ (ours) & $\times$3 
& \textbf{34.85}
 & \textbf{0.9305}
  & \textbf{30.76}
   & \textbf{0.8494}
    & \textbf{29.39}
     & \textbf{0.8122}
      & \textbf{29.31}
       & \textbf{0.8736}
        & \textbf{34.76}
         & \textbf{0.9513}
         
\\
\hline
\hline
Bicubic & $\times$4 
& 28.42
 & 0.8104
  & 26.00
   & 0.7027
    & 25.96
     & 0.6675
      & 23.14
       & 0.6577
        & 24.89
         & 0.7866
                  
\\
SRCNN~\cite{dong2016image} & $\times$4 
& 30.48
 & 0.8628
  & 27.50
   & 0.7513
    & 26.90
     & 0.7101
      & 24.52
       & 0.7221
        & 27.58
         & 0.8555
                   
\\
FSRCNN~\cite{dong2016accelerating} & $\times$4 
& 30.72
 & 0.8660
  & 27.61
   & 0.7550
    & 26.98
     & 0.7150
      & 24.62
       & 0.7280
        & 27.90
         & 0.8610
                   
\\
VDSR~\cite{kim2016accurate} & $\times$4 
& 31.35
 & 0.8830
  & 28.02
   & 0.7680
    & 27.29
     & 0.0726
      & 25.18
       & 0.7540
        & 28.83
         & 0.8870
                   
\\
LapSRN~\cite{lai2017deep} & $\times$4 
& 31.54
 & 0.8850
  & 28.19
   & 0.7720
    & 27.32
     & 0.7270
      & 25.21
       & 0.7560
        & 29.09
         & 0.8900
                   
\\
MemNet~\cite{tai2017memnet} & $\times$4 
& 31.74
 & 0.8893
  & 28.26
   & 0.7723
    & 27.40
     & 0.7281
      & 25.50
       & 0.7630
        & 29.42
         & 0.8942
                   
\\
EDSR~\cite{lim2017enhanced} & $\times$4 
& 32.46
 & 0.8968
  & 28.80
   & 0.7876
    & 27.71
     & 0.7420
      & 26.64
       & 0.8033
        & 31.02
         & 0.9148
                   
\\
SRMDNF~\cite{zhang2018learning} & $\times$4 
& 31.96
 & 0.8925
  & 28.35
   & 0.7787
    & 27.49
     & 0.7337
      & 25.68
       & 0.7731
        & 30.09
         & 0.9024
                   
\\
D-DBPN~\cite{haris2018deep} & $\times$4 
& 32.47
 & 0.8980
  & 28.82
   & 0.7860
    & 27.72
     & 0.7400
      & 26.38
       & 0.7946
        & 30.91
         & 0.9137
         
\\
RDN~\cite{zhang2018residual} & $\times$4 
& 32.47
 & 0.8990
  & 28.81
   & 0.7871
    & 27.72
     & 0.7419
      & 26.61
       & 0.8028
        & 31.00
         & 0.9151
         
\\
RCAN (ours) & $\times$4 
& \underline{32.63}
 & \underline{0.9002}
  & \underline{28.87}
   & \underline{0.7889}
    & \underline{27.77}
     & \underline{0.7436}
      & \underline{26.82}
       & \underline{0.8087}
        & \underline{31.22}
         & \underline{0.9173}

\\
RCAN+ (ours) & $\times$4 
& \textbf{32.73}
 & \textbf{0.9013}
  & \textbf{28.98}
   & \textbf{0.7910}
    & \textbf{27.85}
     & \textbf{0.7455}
      & \textbf{27.10}
       & \textbf{0.8142}
        & \textbf{31.65}
         & \textbf{0.9208}
              
\\
\hline
\hline
Bicubic & $\times$8 
& 24.40
 & 0.6580
  & 23.10
   & 0.5660
    & 23.67
     & 0.5480
      & 20.74
       & 0.5160
        & 21.47
         & 0.6500
                 
\\
SRCNN~\cite{dong2016image} & $\times$8 
& 25.33
 & 0.6900
  & 23.76
   & 0.5910
    & 24.13
     & 0.5660
      & 21.29
       & 0.5440
        & 22.46
         & 0.6950
                   
\\
FSRCNN~\cite{dong2016accelerating} & $\times$8 
& 20.13
 & 0.5520
  & 19.75
   & 0.4820
    & 24.21
     & 0.5680
      & 21.32
       & 0.5380
        & 22.39
         & 0.6730
                   
\\
SCN~\cite{wang2015deep} & $\times$8 
& 25.59
 & 0.7071
  & 24.02
   & 0.6028
    & 24.30
     & 0.5698
      & 21.52
       & 0.5571
        & 22.68
         & 0.6963

\\
VDSR~\cite{kim2016accurate} & $\times$8 
& 25.93
 & 0.7240
  & 24.26
   & 0.6140
    & 24.49
     & 0.5830
      & 21.70
       & 0.5710
        & 23.16
         & 0.7250
                   
\\   
LapSRN~\cite{lai2017deep} & $\times$8 
& 26.15
 & 0.7380
  & 24.35
   & 0.6200
    & 24.54
     & 0.5860
      & 21.81
       & 0.5810
        & 23.39
         & 0.7350
                   
\\
MemNet~\cite{tai2017memnet} & $\times$8 
& 26.16
 & 0.7414
  & 24.38
   & 0.6199
    & 24.58
     & 0.5842
      & 21.89
       & 0.5825
        & 23.56
         & 0.7387

\\
MSLapSRN~\cite{MSLapSRN} & $\times$8 
& 26.34
 & 0.7558
  & 24.57
   & 0.6273
    & 24.65
     & 0.5895
      & 22.06
       & 0.5963
        & 23.90
         & 0.7564
                   
\\
EDSR~\cite{lim2017enhanced} & $\times$8 
& 26.96
 & 0.7762
  & 24.91
   & 0.6420
    & 24.81
     & 0.5985
      & 22.51
       & 0.6221
        & 24.69
         & 0.7841
                   
\\
D-DBPN~\cite{haris2018deep} & $\times$8 
& 27.21
 & 0.7840
  & 25.13
   & 0.6480
    & 24.88
     & 0.6010
      & 22.73
       & 0.6312
        & 25.14
         & 0.7987
         
\\
RCAN (ours) & $\times$8 
& \underline{27.31}
 & \underline{0.7878}
  & \underline{25.23}
   & \underline{0.6511}
    & \underline{24.98}
     & \underline{0.6058}
      & \underline{23.00}
       & \underline{0.6452}
        & \underline{25.24}
         & \underline{0.8029}

\\
RCAN+ (ours) & $\times$8 
& \textbf{27.47}
 & \textbf{0.7913}
  & \textbf{25.40}
   & \textbf{0.6553}
    & \textbf{25.05}
     & \textbf{0.6077}
      & \textbf{23.22}
       & \textbf{0.6524}
        & \textbf{25.58}
         & \textbf{0.8092}
           
\\
\hline             
\end{tabular}
\end{center}
\vspace{-5mm}
\end{table}

\vspace{-3mm}
\begin{figure}[thbp]
	\newlength\fsdurthree
	\setlength{\fsdurthree}{-1.5mm}
	\scriptsize
	\centering
	\begin{tabular}{cc}
		\begin{adjustbox}{valign=t}
		\tiny
			\begin{tabular}{c}
				\includegraphics[width=0.229\textwidth]{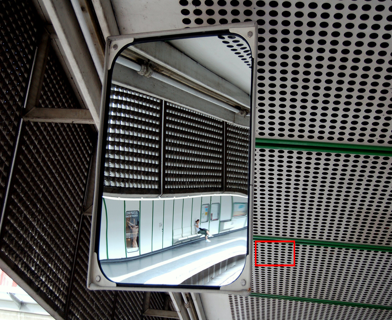}
				\\
				Urban100 ($4\times$):
				\\
				img\_004
			\end{tabular}
		\end{adjustbox}
		\hspace{-2.3mm}
		\begin{adjustbox}{valign=t}
		\tiny
			\begin{tabular}{cccccc}
				\includegraphics[width=\widthscalefive \textwidth]{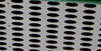} \hspace{\fsdurthree} &
				\includegraphics[width=\widthscalefive \textwidth]{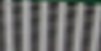} \hspace{\fsdurthree} &
				\includegraphics[width=\widthscalefive \textwidth]{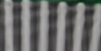} \hspace{\fsdurthree} &
				\includegraphics[width=\widthscalefive \textwidth]{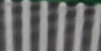} \hspace{\fsdurthree} &
				\includegraphics[width=\widthscalefive \textwidth]{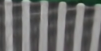} 
				\\
				HR \hspace{\fsdurthree} &
				Bicubic \hspace{\fsdurthree} &
				SRCNN~\cite{dong2016image} \hspace{\fsdurthree} &
				FSRCNN~\cite{dong2016accelerating} \hspace{\fsdurthree} &
				VDSR~\cite{kim2016accurate} 
				\\
				PSNR/SSIM \hspace{\fsdurthree} &
				21.08/0.6788 \hspace{\fsdurthree} &
				22.13/0.7635 \hspace{\fsdurthree} &
				22.02/0.7628 \hspace{\fsdurthree} &
				22.37/0.7939 
				\\
				\includegraphics[width=\widthscalefive \textwidth]{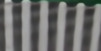} \hspace{\fsdurthree} &
				\includegraphics[width=\widthscalefive \textwidth]{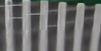} \hspace{\fsdurthree} &
				\includegraphics[width=\widthscalefive \textwidth]{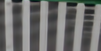} \hspace{\fsdurthree} &
				\includegraphics[width=\widthscalefive \textwidth]{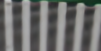} \hspace{\fsdurthree} &
				\includegraphics[width=\widthscalefive \textwidth]{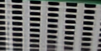}  
				\\ 
				LapSRN~\cite{lai2017deep} \hspace{\fsdurthree} &
				MemNet~\cite{tai2017memnet} \hspace{\fsdurthree} &
				EDSR~\cite{lim2017enhanced} \hspace{\fsdurthree} &
				SRMDNF~\cite{zhang2018learning} \hspace{\fsdurthree} &
				RCAN 
				\\
				22.41/0.7984 \hspace{\fsdurthree} &
				22.35/0.7992 \hspace{\fsdurthree} &
				24.07/0.8591 \hspace{\fsdurthree} &
				22.93/0.8207 \hspace{\fsdurthree} &
				\textbf{25.64}/\textbf{0.8830} \hspace{\fsdurthree} 
				\\
			\end{tabular}
		\end{adjustbox}
		\vspace{0.5mm}
		\\
		\begin{adjustbox}{valign=t}
		\tiny
			\begin{tabular}{c}
				\includegraphics[width=0.229\textwidth]{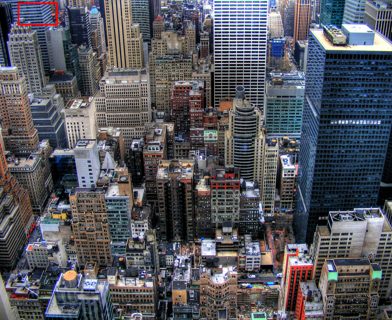}
				\\
				Urban100 ($4\times$):
				\\
				img\_073
			\end{tabular}
		\end{adjustbox}
		\hspace{-2.3mm}
		\begin{adjustbox}{valign=t}
		\tiny
			\begin{tabular}{cccccc}
				\includegraphics[width=\widthscalefive \textwidth]{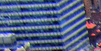} \hspace{\fsdurthree} &
				\includegraphics[width=\widthscalefive \textwidth]{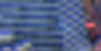} \hspace{\fsdurthree} &
				\includegraphics[width=\widthscalefive \textwidth]{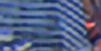} \hspace{\fsdurthree} &
				\includegraphics[width=\widthscalefive \textwidth]{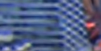} \hspace{\fsdurthree} &
				\includegraphics[width=\widthscalefive \textwidth]{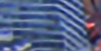} 
				\\
				HR \hspace{\fsdurthree} &
				Bicubic \hspace{\fsdurthree} &
				SRCNN~\cite{dong2016image} \hspace{\fsdurthree} &
				FSRCNN~\cite{dong2016accelerating} \hspace{\fsdurthree} &
				VDSR~\cite{kim2016accurate} 
				\\
				PSNR/SSIM \hspace{\fsdurthree} &
				19.48/0.4371 \hspace{\fsdurthree} &
				19.94/0.5124 \hspace{\fsdurthree} &
				19.88/0.5158 \hspace{\fsdurthree} &
				19.88/0.5229 
				\\
				\includegraphics[width=\widthscalefive \textwidth]{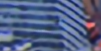} \hspace{\fsdurthree} &
				\includegraphics[width=\widthscalefive \textwidth]{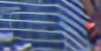} \hspace{\fsdurthree} &
				\includegraphics[width=\widthscalefive \textwidth]{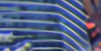} \hspace{\fsdurthree} &
				\includegraphics[width=\widthscalefive \textwidth]{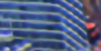} \hspace{\fsdurthree} &
				\includegraphics[width=\widthscalefive \textwidth]{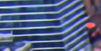}  
				\\ 
				LapSRN~\cite{lai2017deep} \hspace{\fsdurthree} &
				MemNet~\cite{tai2017memnet} \hspace{\fsdurthree} &
				EDSR~\cite{lim2017enhanced} \hspace{\fsdurthree} &
				SRMDNF~\cite{zhang2018learning} \hspace{\fsdurthree} &
				RCAN 
				\\
				19.76/0.5250 \hspace{\fsdurthree} &
				19.71/0.5213 \hspace{\fsdurthree} &
				20.42/0.6028 \hspace{\fsdurthree} &
				19.88/0.5425 \hspace{\fsdurthree} &
				\textbf{21.26}/\textbf{0.6298} \hspace{\fsdurthree} 
				\\
			\end{tabular}
		\end{adjustbox}
		\vspace{0.5mm}
		\\		
		\begin{adjustbox}{valign=t}
		\tiny
			\begin{tabular}{c}
				\includegraphics[width=0.229\textwidth]{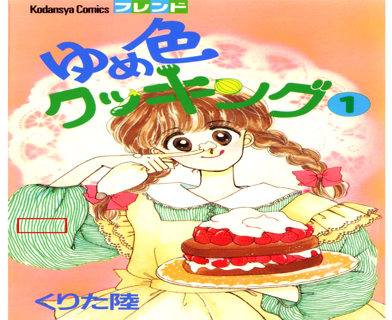}
				\\
				Manga109 ($4\times$):
				\\
				YumeiroCooking
			\end{tabular}
		\end{adjustbox}
		\hspace{-2.3mm}
		\begin{adjustbox}{valign=t}
		\tiny
			\begin{tabular}{cccccc}
				\includegraphics[width=\widthscalefive \textwidth]{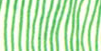} \hspace{\fsdurthree} &
				\includegraphics[width=\widthscalefive \textwidth]{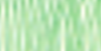} \hspace{\fsdurthree} &
				\includegraphics[width=\widthscalefive \textwidth]{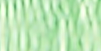} \hspace{\fsdurthree} &
				\includegraphics[width=\widthscalefive \textwidth]{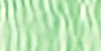} \hspace{\fsdurthree} &
				\includegraphics[width=\widthscalefive \textwidth]{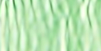} 
				\\
				HR \hspace{\fsdurthree} &
				Bicubic \hspace{\fsdurthree} &
				SRCNN~\cite{dong2016image} \hspace{\fsdurthree} &
				FSRCNN~\cite{dong2016accelerating} \hspace{\fsdurthree} &
				VDSR~\cite{kim2016accurate} 
				\\
				PSNR/SSIM \hspace{\fsdurthree} &
				24.66/0.7849 \hspace{\fsdurthree} &
				26.22/0.8464 \hspace{\fsdurthree} &
				26.38/0.8496 \hspace{\fsdurthree} &
				26.89/0.8703 
				\\
				\includegraphics[width=\widthscalefive \textwidth]{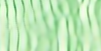} \hspace{\fsdurthree} &
				\includegraphics[width=\widthscalefive \textwidth]{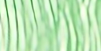} \hspace{\fsdurthree} &
				\includegraphics[width=\widthscalefive \textwidth]{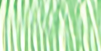} \hspace{\fsdurthree} &
				\includegraphics[width=\widthscalefive \textwidth]{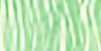} \hspace{\fsdurthree} &
				\includegraphics[width=\widthscalefive \textwidth]{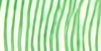}  
				\\ 
				LapSRN~\cite{lai2017deep} \hspace{\fsdurthree} &
				MemNet~\cite{tai2017memnet} \hspace{\fsdurthree} &
				EDSR~\cite{lim2017enhanced} \hspace{\fsdurthree} &
				SRMDNF~\cite{zhang2018learning} \hspace{\fsdurthree} &
				RCAN 
				\\
				26.92/0.8739 \hspace{\fsdurthree} &
				27.09/0.8811 \hspace{\fsdurthree} &
				29.04/0.9230 \hspace{\fsdurthree} &
				27.53/0.8901 \hspace{\fsdurthree} &
				\textbf{29.85}/\textbf{0.9368} \hspace{\fsdurthree} 
				\\
			\end{tabular}
		\end{adjustbox}
		\vspace{-3mm}
	\end{tabular}
	\caption{
		Visual comparison for $4\times$ SR with BI model on Urban100 and Manga109 datasets. The best results are \textbf{highlighted}
	}
\label{fig:result_4x_Urban100_Manga109}
\vspace{-5mm}
\end{figure}

\vspace{-4mm}
\subsection{Results with Bicubic (BI) Degradation Model}
\label{subsec:results_BI}
\vspace{-2mm}
We compare our method with 11 state-of-the-art methods: SRCNN~\cite{dong2016image}, FSRCNN~\cite{dong2016accelerating}, SCN~\cite{wang2015deep}, VDSR~\cite{kim2016accurate}, LapSRN~\cite{lai2017deep}, MemNet~\cite{tai2017memnet}, EDSR~\cite{lim2017enhanced}, SRMDNF~\cite{zhang2018learning}, D-DBPN~\cite{haris2018deep}, and RDN~\cite{zhang2018residual}. Similar to~\cite{timofte2016seven,lim2017enhanced,zhang2018residual}, we also introduce self-ensemble strategy to further improve our RCAN and denote the self-ensembled one as RCAN+. More comparisons are provided in supplementary material.

\textbf{Quantitative results by PSNR/SSIM.} Table~\ref{tab:results_psnr_ssim_x2348} shows quantitative comparisons for $\times$2, $\times$3, $\times$4, and $\times$8 SR. The results of D-DBPN~\cite{haris2018deep} are cited from their paper. When compared with all previous methods, our RCAN+ performs the best on all the datasets with all scaling factors. Even without self-ensemble, our RCAN also outperforms other compared methods. 

On the other hand, when the scaling factor become larger (e.g., 8), the gains of our RCAN over EDSR also becomes larger. For Urban100 and Manga109, the PSNR gains of RCAN over EDSR are 0.49 dB and 0.55 dB. EDSR has much larger number of parameters (43 M) than ours (16 M), but our RCAN obtains much better performance. Instead of constantly rescaling the features in EDSR, our RCAN adaptively rescales features with channel attention (CA). CA allows our network to further focus on more informative features. This observation indicates that very large network depth and CA improve the performance. 
 

\begin{figure}[thbp]
	\newlength\fsdttwofig
	\setlength{\fsdttwofig}{-1.5mm}
	\scriptsize
	\centering
	\begin{tabular}{cc}
		\begin{adjustbox}{valign=t}
		\tiny
			\begin{tabular}{c}
				\includegraphics[width=0.229\textwidth]{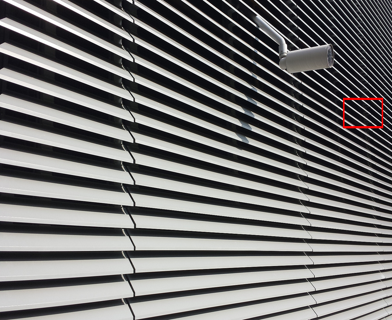}
				\\
				 Urban100 ($8\times$):
				\\
				img\_040
				
			\end{tabular}
		\end{adjustbox}
		\hspace{-2.3mm}
		\begin{adjustbox}{valign=t}
		\tiny
			\begin{tabular}{cccccc}
				\includegraphics[width=\widthscalefive \textwidth]{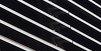} \hspace{\fsdttwofig} &
				\includegraphics[width=\widthscalefive \textwidth]{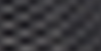} \hspace{\fsdttwofig} &
				\includegraphics[width=\widthscalefive \textwidth]{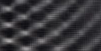} \hspace{\fsdttwofig} &
				\includegraphics[width=\widthscalefive \textwidth]{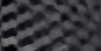} \hspace{\fsdttwofig} &
				\includegraphics[width=\widthscalefive \textwidth]{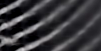} 
				\\
				HR \hspace{\fsdttwofig} &
				Bicubic \hspace{\fsdttwofig} &
				SRCNN~\cite{dong2016image} \hspace{\fsdttwofig} &
				SCN~\cite{wang2015deep} \hspace{\fsdttwofig} &
				VDSR~\cite{kim2016accurate}
				\\
				PSNR/SSIM \hspace{\fsdttwofig} &
				15.89/0.4595 \hspace{\fsdttwofig} &
				17.48/0.5927 \hspace{\fsdttwofig} &
				17.64/0.6410 \hspace{\fsdttwofig} &
				17.59/0.6612 
				\\
				\includegraphics[width=\widthscalefive \textwidth]{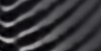} \hspace{\fsdttwofig} &
				\includegraphics[width=\widthscalefive \textwidth]{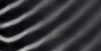} \hspace{\fsdttwofig} &
				\includegraphics[width=\widthscalefive \textwidth]{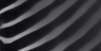} \hspace{\fsdttwofig} &
				\includegraphics[width=\widthscalefive \textwidth]{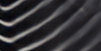} \hspace{\fsdttwofig} &
				\includegraphics[width=\widthscalefive \textwidth]{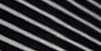}  
				\\ 
				LapSRN~\cite{lai2017deep} \hspace{\fsdttwofig} &
				MemNet~\cite{tai2017memnet} \hspace{\fsdttwofig} &
				MSLapSRN~\cite{MSLapSRN} \hspace{\fsdttwofig} &
				EDSR~\cite{lim2017enhanced}  \hspace{\fsdttwofig} &
				RCAN 
				\\
				18.27/0.7182 \hspace{\fsdttwofig} &
				18.17/0.7190 \hspace{\fsdttwofig} &
				18.52/0.7525 \hspace{\fsdttwofig} &
				19.53/0.7857 \hspace{\fsdttwofig} &
				\textbf{22.43}/\textbf{0.8607} \hspace{\fsdttwofig} 
				\\
			\end{tabular}
		\end{adjustbox}
		\vspace{0.5mm}
		\\
		\begin{adjustbox}{valign=t}
		\tiny
			\begin{tabular}{c}
				\includegraphics[width=0.229\textwidth]{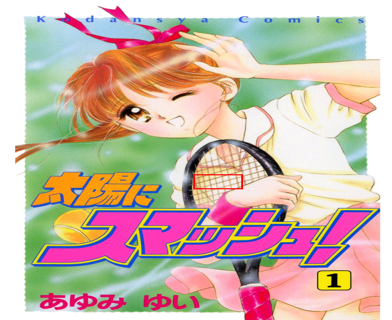}
				\\
				Manga109 ($8\times$):
				\\
				TaiyouNiSmash
			\end{tabular}
		\end{adjustbox}
		\hspace{-2.3mm}
		\begin{adjustbox}{valign=t}
		\tiny
			\begin{tabular}{cccccc}
				\includegraphics[width=\widthscalefive \textwidth]{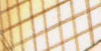} \hspace{\fsdttwofig} &
				\includegraphics[width=\widthscalefive \textwidth]{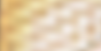} \hspace{\fsdttwofig} &
				\includegraphics[width=\widthscalefive \textwidth]{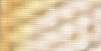} \hspace{\fsdttwofig} &
				\includegraphics[width=\widthscalefive \textwidth]{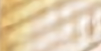} \hspace{\fsdttwofig} &
				\includegraphics[width=\widthscalefive \textwidth]{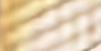} 
				\\
				HR \hspace{\fsdttwofig} &
				Bicubic \hspace{\fsdttwofig} &
				SRCNN~\cite{dong2016image} \hspace{\fsdttwofig} &
				SCN~\cite{wang2015deep} \hspace{\fsdttwofig} &
				VDSR~\cite{kim2016accurate} 
				\\
				PSNR/SSIM \hspace{\fsdttwofig} &
				24.89/0.7572 \hspace{\fsdttwofig} &
				25.58/0.6993 \hspace{\fsdttwofig} &
				26.62/0.8035 \hspace{\fsdttwofig} &
				26.33/0.8091 
				\\
				\includegraphics[width=\widthscalefive \textwidth]{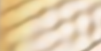} \hspace{\fsdttwofig} &
				\includegraphics[width=\widthscalefive \textwidth]{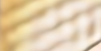} \hspace{\fsdttwofig} &
				\includegraphics[width=\widthscalefive \textwidth]{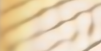} \hspace{\fsdttwofig} &
				\includegraphics[width=\widthscalefive \textwidth]{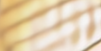} \hspace{\fsdttwofig} &
				\includegraphics[width=\widthscalefive \textwidth]{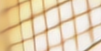}  
				\\ 
				LapSRN~\cite{lai2017deep} \hspace{\fsdttwofig} &
				MemNet~\cite{tai2017memnet} \hspace{\fsdttwofig} &
				MSLapSRN~\cite{MSLapSRN} \hspace{\fsdttwofig} &
				EDSR~\cite{lim2017enhanced}  \hspace{\fsdttwofig} &
				RCAN  
				\\
				27.26/0.8278 \hspace{\fsdttwofig} &
				27.47/0.8353 \hspace{\fsdttwofig} &
				28.02/0.8532 \hspace{\fsdttwofig} &
				29.44/0.8746 \hspace{\fsdttwofig} &
				\textbf{30.67}/\textbf{0.8961} \hspace{\fsdttwofig} 
				\\
			\end{tabular}
		\end{adjustbox}
		\vspace{-2mm}
	\end{tabular}
	\caption{
		Visual comparison for $8\times$ SR with BI model on Urban100 and Manga109 datasets. The best results are \textbf{highlighted}
	}
\label{fig:result_8x}
\vspace{-8mm}
\end{figure}

\textbf{Visual results.} In Figure~\ref{fig:result_4x_Urban100_Manga109}, we show visual comparisons on scale $\times$4. For image ``img\_004", we observe that most of the compared methods cannot recover the lattices and would suffer from blurring artifacts. In contrast, our RCAN can alleviate the blurring artifacts better and recover more details. For image ``img\_073", most of the compared methods produce blurring artifacts along the horizontal lines. What's worse, for the right parts of the cropped images, FSRCNN cannot recover lines. Other methods would generate some lines with wrong directions. Only our RCAN produces more faithful results.  For image ``YumeiroCooking", the cropped part is full of textures. As we can see, all the compared methods suffer from heavy blurring artifacts, failing to recover more details. While, our RCAN can recover them obviously, being more faithful to the ground truth. Such obvious comparisons demonstrate that networks with more powerful representational ability can extract more sophisticated features from the LR space. 

To further illustrate the analyses above, we show visual comparisons for 8$\times$ SR in Figure~\ref{fig:result_8x}. For image ``img\_040", due to very large scaling factor, the result by Bicubic would lose the structures and produce different structures. This wrong pre-scaling result would also lead some state-of-the-art methods (e.g., SRCNN, VDSR, and MemNet) to generate totally wrong structures. Even starting from the original LR input, other methods cannot recover the right structure either. While, our RCAN can recover them correctly. For smaller details, like the net in image ``TaiyouNiSmash", the tiny lines can be lost in the LR image. When the scaling factor is very large (e.g., 8), LR images contain very limited information for SR. Losing most high-frequency information makes it very difficult for SR methods to reconstruct informative results. Most of compared methods cannot achieve this goal and produce serious blurring artifacts. However, our RCAN can obtain more useful information and produce finer results. 

As we have discussed above, in BI degradation model, the reconstruction of high-frequency information is very important and difficult, especially with large scaling factor (e.g., 8). Our proposed RIR structure makes the main network learn residual information. Channel attention (CA) is further used to enhance the representational ability of the network by adaptively rescaling channel-wise features. 

\begin{table}[htbp]
\scriptsize
\center
\begin{center}
\caption{Quantitative results with BD degradation model. Best and second best results are \textbf{highlighted} and \underline{underlined}}
\label{tab:results_psnr_ssim_BD}
\vspace{-3mm}
\begin{tabular}{|l|c|c|c|c|c|c|c|c|c|c|c|}
\hline
\multirow{2}{*}{Method} & \multirow{2}{*}{Scale} &  \multicolumn{2}{c|}{Set5} &  \multicolumn{2}{c|}{Set14} &  \multicolumn{2}{c|}{B100} &  \multicolumn{2}{c|}{Urban100} &  \multicolumn{2}{c|}{Manga109}  
\\
\cline{3-12}
&  & PSNR & SSIM & PSNR & SSIM & PSNR & SSIM & PSNR & SSIM & PSNR & SSIM 
\\
\hline
\hline
Bicubic & $\times$3 
& 28.78
 & 0.8308
  & 26.38
   & 0.7271
    & 26.33
     & 0.6918
      & 23.52
       & 0.6862
        & 25.46
         & 0.8149 
\\
SPMSR~\cite{peleg2014statistical} & $\times$3 
& 32.21
 & 0.9001
  & 28.89
   & 0.8105
    & 28.13
     & 0.7740
      & 25.84
       & 0.7856
        & 29.64
         & 0.9003
          
\\
SRCNN~\cite{dong2016image} & $\times$3
& 32.05
 & 0.8944
  & 28.80
   & 0.8074
    & 28.13
     & 0.7736
      & 25.70
       & 0.7770
        & 29.47
         & 0.8924
          
\\
FSRCNN~\cite{dong2016accelerating} & $\times$3 
& 26.23
 & 0.8124
  & 24.44
   & 0.7106
    & 24.86
     & 0.6832
      & 22.04
       & 0.6745
        & 23.04
         & 0.7927
          
\\
VDSR~\cite{kim2016accurate} & $\times$3 
& 33.25
 & 0.9150
  & 29.46
   & 0.8244
    & 28.57
     & 0.7893
      & 26.61
       & 0.8136
        & 31.06
         & 0.9234
          
\\
IRCNN~\cite{zhang2017learning} & $\times$3 
& 33.38
 & 0.9182
  & 29.63
   & 0.8281
    & 28.65
     & 0.7922
      & 26.77
       & 0.8154
        & 31.15
         & 0.9245
          
\\
SRMDNF~\cite{zhang2018learning} & $\times$3 
& 34.01
 & 0.9242
  & 30.11
   & 0.8364
    & 28.98
     & 0.8009
      & 27.50
       & 0.8370
        & 32.97
         & 0.9391
          
\\
RDN~\cite{zhang2018residual} & $\times$3 
& 34.58
 & 0.9280
  & 30.53
   & 0.8447
    & 29.23
     & 0.8079
      & 28.46
       & 0.8582
        & 33.97
         & 0.9465
          
\\
RCAN (ours) & $\times$3 
& \underline{34.70}
 & \underline{0.9288}
  & \underline{30.63}
   & \underline{0.8462}
    & \underline{29.32}
     & \underline{0.8093}
      & \underline{28.81}
       & \underline{0.8647}
        & \underline{34.38}
         & \underline{0.9483}
          
\\
RCAN+ (ours) & $\times$3 
& \textbf{34.83}
 & \textbf{0.9296}
  & \textbf{30.76}
   & \textbf{0.8479}
    & \textbf{29.39}
     & \textbf{0.8106}
      & \textbf{29.04}
       & \textbf{0.8682}
        & \textbf{34.76}
         & \textbf{0.9502}
          
\\
\hline             
\end{tabular}
\end{center}
\vspace{-4mm}
\end{table}

\begin{figure}[htbp]
	\newlength\fsdttwofigBD
	\setlength{\fsdttwofigBD}{-1.5mm}
	\scriptsize
	\centering
	\begin{tabular}{cc}
		\begin{adjustbox}{valign=t}
		\tiny
			\begin{tabular}{c}
				\includegraphics[width=0.229\textwidth]{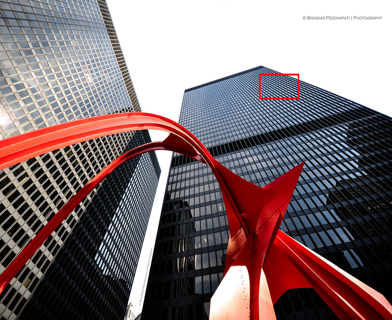}
				\\
				 Urban100 ($3\times$):
				\\
				img\_062
				
			\end{tabular}
		\end{adjustbox}
		\hspace{-2.3mm}
		\begin{adjustbox}{valign=t}
		\tiny
			\begin{tabular}{cccccc}
				\includegraphics[width=\widthscalefive \textwidth]{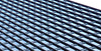} \hspace{\fsdttwofigBD} &
				\includegraphics[width=\widthscalefive \textwidth]{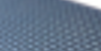} \hspace{\fsdttwofigBD} &
				\includegraphics[width=\widthscalefive \textwidth]{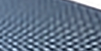} \hspace{\fsdttwofigBD} &
				\includegraphics[width=\widthscalefive \textwidth]{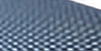} \hspace{\fsdttwofigBD} &
				\includegraphics[width=\widthscalefive \textwidth]{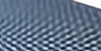} 
				\\
				HR \hspace{\fsdttwofigBD} &
				Bicubic \hspace{\fsdttwofigBD} &
				SPMSR~\cite{peleg2014statistical} \hspace{\fsdttwofigBD} &
				SRCNN~\cite{dong2016image} \hspace{\fsdttwofigBD} &
				FSRCNN~\cite{dong2016accelerating}
				\\
				PSNR/SSIM \hspace{\fsdttwofigBD} &
				20.20/0.6737 \hspace{\fsdttwofigBD} &
				21.72/0.7923 \hspace{\fsdttwofigBD} &
				21.74/0.7882 \hspace{\fsdttwofigBD} &
				19.30/0.6960 
				\\
				\includegraphics[width=\widthscalefive \textwidth]{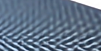} \hspace{\fsdttwofigBD} &
				\includegraphics[width=\widthscalefive \textwidth]{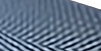} \hspace{\fsdttwofigBD} &
				\includegraphics[width=\widthscalefive \textwidth]{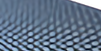} \hspace{\fsdttwofigBD} &
				\includegraphics[width=\widthscalefive \textwidth]{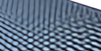} \hspace{\fsdttwofigBD} &
				\includegraphics[width=\widthscalefive \textwidth]{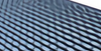}  
				\\ 
				VDSR~\cite{kim2016accurate} \hspace{\fsdttwofigBD} &
				IRCNN~\cite{zhang2017learning} \hspace{\fsdttwofigBD} &
				SRMDNF~\cite{zhang2018learning} \hspace{\fsdttwofigBD} &
				RDN~\cite{zhang2018residual}  \hspace{\fsdttwofigBD} &
				RCAN 
				\\
				22.36/0.8351 \hspace{\fsdttwofigBD} &
				22.32/0.8292 \hspace{\fsdttwofigBD} &
				23.11/0.8662 \hspace{\fsdttwofigBD} &
				24.42/0.9052 \hspace{\fsdttwofigBD} &
				\textbf{25.73}/\textbf{0.9238} \hspace{\fsdttwofigBD} 
				\\
			\end{tabular}
		\end{adjustbox}
		\vspace{0.5mm}
		\\
		\begin{adjustbox}{valign=t}
		\tiny
			\begin{tabular}{c}
				\includegraphics[width=0.229\textwidth]{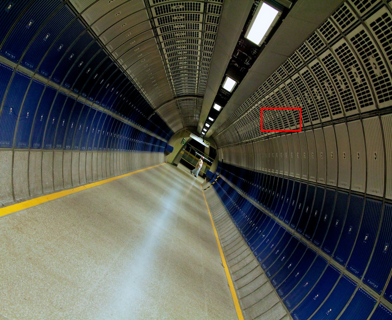}
				\\
				Urban100 ($3\times$):
				\\
				img\_078
			\end{tabular}
		\end{adjustbox}
		\hspace{-2.3mm}
		\begin{adjustbox}{valign=t}
		\tiny
			\begin{tabular}{cccccc}
				\includegraphics[width=\widthscalefive \textwidth]{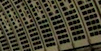} \hspace{\fsdttwofigBD} &
				\includegraphics[width=\widthscalefive \textwidth]{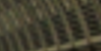} \hspace{\fsdttwofigBD} &
				\includegraphics[width=\widthscalefive \textwidth]{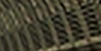} \hspace{\fsdttwofigBD} &
				\includegraphics[width=\widthscalefive \textwidth]{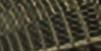} \hspace{\fsdttwofigBD} &
				\includegraphics[width=\widthscalefive \textwidth]{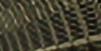} 
				\\
				HR \hspace{\fsdttwofigBD} &
				Bicubic \hspace{\fsdttwofigBD} &
				SPMSR~\cite{peleg2014statistical} \hspace{\fsdttwofigBD} &
				SRCNN~\cite{dong2016image} \hspace{\fsdttwofigBD} &
				FSRCNN~\cite{dong2016accelerating}
				\\
				PSNR/SSIM \hspace{\fsdttwofigBD} &
				26.10/0.7032 \hspace{\fsdttwofigBD} &
				28.06/0.7950 \hspace{\fsdttwofigBD} &
				27.91/0.7874 \hspace{\fsdttwofigBD} &
				24.34/0.6711 
				\\
				\includegraphics[width=\widthscalefive \textwidth]{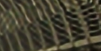} \hspace{\fsdttwofigBD} &
				\includegraphics[width=\widthscalefive \textwidth]{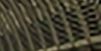} \hspace{\fsdttwofigBD} &
				\includegraphics[width=\widthscalefive \textwidth]{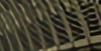} \hspace{\fsdttwofigBD} &
				\includegraphics[width=\widthscalefive \textwidth]{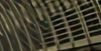} \hspace{\fsdttwofigBD} &
				\includegraphics[width=\widthscalefive \textwidth]{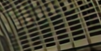}  
				\\ 
				VDSR~\cite{kim2016accurate} \hspace{\fsdttwofigBD} &
				IRCNN~\cite{zhang2017learning} \hspace{\fsdttwofigBD} &
				SRMDNF~\cite{zhang2018learning} \hspace{\fsdttwofigBD} &
				RDN~\cite{zhang2018residual}  \hspace{\fsdttwofigBD} &
				RCAN
				\\
				28.34/0.8166 \hspace{\fsdttwofigBD} &
				28.57/0.8184 \hspace{\fsdttwofigBD} &
				29.08/0.8342 \hspace{\fsdttwofigBD} &
				29.94/0.8513 \hspace{\fsdttwofigBD} &
				\textbf{30.65}/\textbf{0.8624} \hspace{\fsdttwofigBD} 
				\\
			\end{tabular}
		\end{adjustbox}
		\vspace{-3mm}
	\end{tabular}
	\caption{
		Visual comparison for $3\times$ SR with BD model on Urban100 dataset. The best results are \textbf{highlighted}
	}
	\label{fig:result_BD3x}
\vspace{-9mm}
\end{figure}

\subsection{Results with Blur-downscale (BD) Degradation Model} 
\vspace{-7mm}
We further apply our method to super-resolve images with blur-down (BD) degradation model, which is also commonly used recently~\cite{zhang2017learning,zhang2018learning,zhang2018residual}. 

\textbf{Quantitative results by PSNR/SSIM.} Here, we compare 3$\times$ SR results with 7 state-of-the-art methods: SPMSR~\cite{peleg2014statistical}, SRCNN~\cite{dong2016image}, FSRCNN~\cite{dong2016accelerating}, VDSR~\cite{kim2016accurate}, IRCNN~\cite{zhang2017learning}, SRMDNF~\cite{zhang2018learning}, and RDN~\cite{zhang2018residual}. As shown in Table~\ref{tab:results_psnr_ssim_BD}, RDN has achieved very high performance on each dataset. While, our RCAN can obtain notable gains over RDN. Using self-ensemble, RCAN+ achieves even better results. Compared with fully using hierarchical features in RDN, a much deeper network with channel attention in RCAN achieves better performance. This comparison also indicates that there has promising potential to investigate much deeper networks for image SR.

\textbf{Visual Results.} We also show visual comparisons in Figure~\ref{fig:result_BD3x}. For challenging details in images ``img\_062" and ``img\_078", most methods suffer from heavy blurring artifacts. RDN alleviates it to some degree and can recover more details. In contrast, our RCAN obtains much better results by recovering more informative components. These comparisons indicate that very deep channel attention guided network would alleviate the blurring artifacts. It also demonstrates the strong ability of RCAN for BD degradation model.

\begin{table}[htbp]
\centering
\begin{center}

\caption{ResNet object recognition performance. The best results are \textbf{highlighted}}
\label{tab:results_object_recognition}
\vspace{-3mm}
\begin{tabular}{|c|c|c|c|c|c|c|c|c|c|c|c|}
\hline
\multirow{1}{*}{Evaluation} & \multirow{1}{*}{Bicubic} & DRCN~\cite{kim2016deeply} & FSRCNN~\cite{dong2016accelerating} & PSyCo~\cite{perez2016psyco} & ENet-E~\cite{sajjadi2017enhancenet} & RCAN  & \multirow{1}{*}{Baseline}
\\
\hline
Top-1 error & 0.506 & 0.477 & 0.437  & 0.454 & 0.449 &  \textbf{0.393}  & 0.260 
\\
\hline
Top-5 error & 0.266 & 0.242 & 0.196 & 0.224 & 0.214 &  \textbf{0.167} & 0.072 
\\
\hline
\end{tabular}
\end{center}
\vspace{-8mm}
\end{table}

\begin{figure}[htbp]
\scriptsize
\centering
\centerline{
\subfigure[Results on Set5 (4$\times$)]{
\label{fig:investeD}
\includegraphics[width = 60mm, height = 30mm]{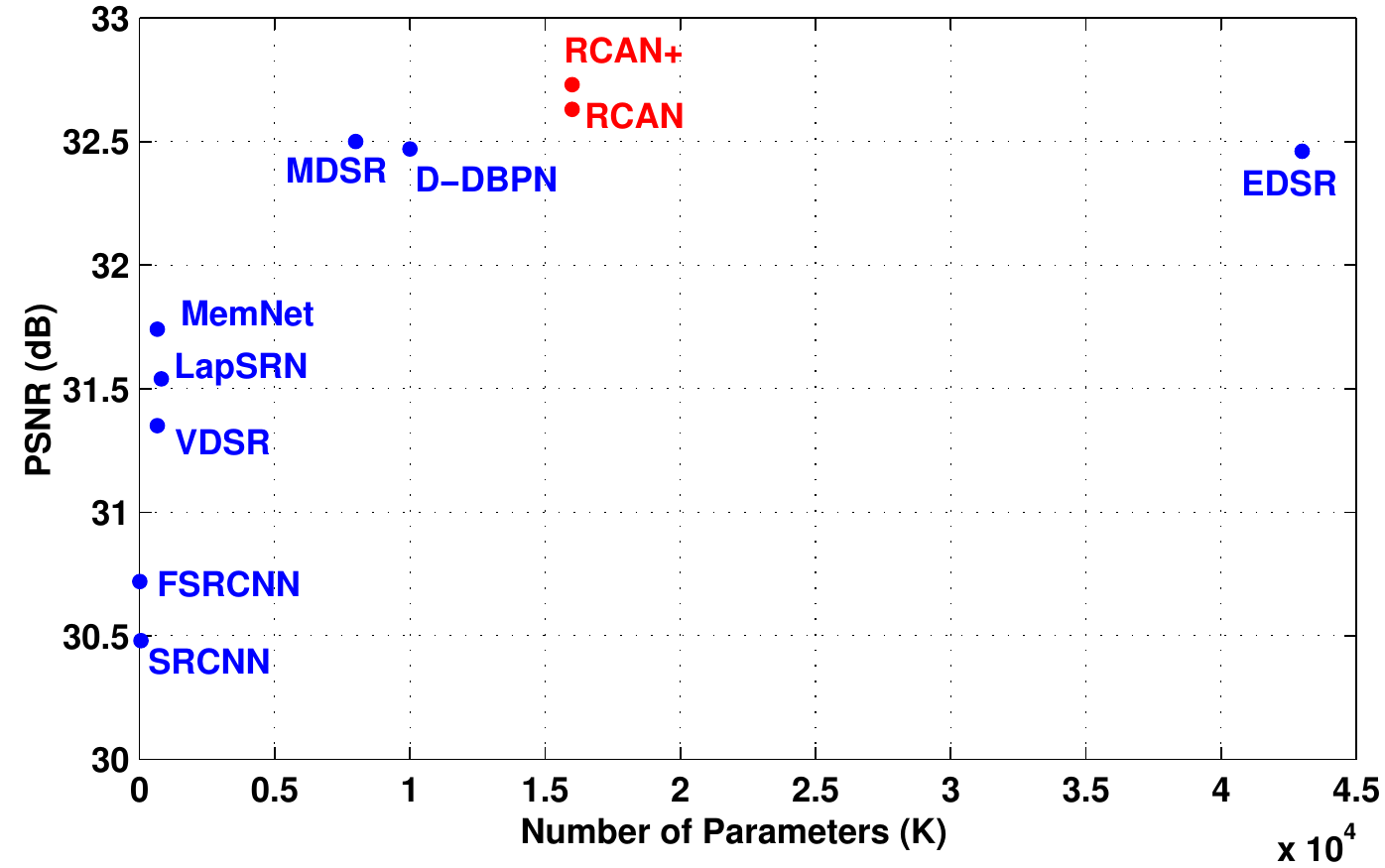}}
\subfigure[Results on Set5 (8$\times$)]{
\label{fig:investeC}
\includegraphics[width = 60mm, height = 30mm]{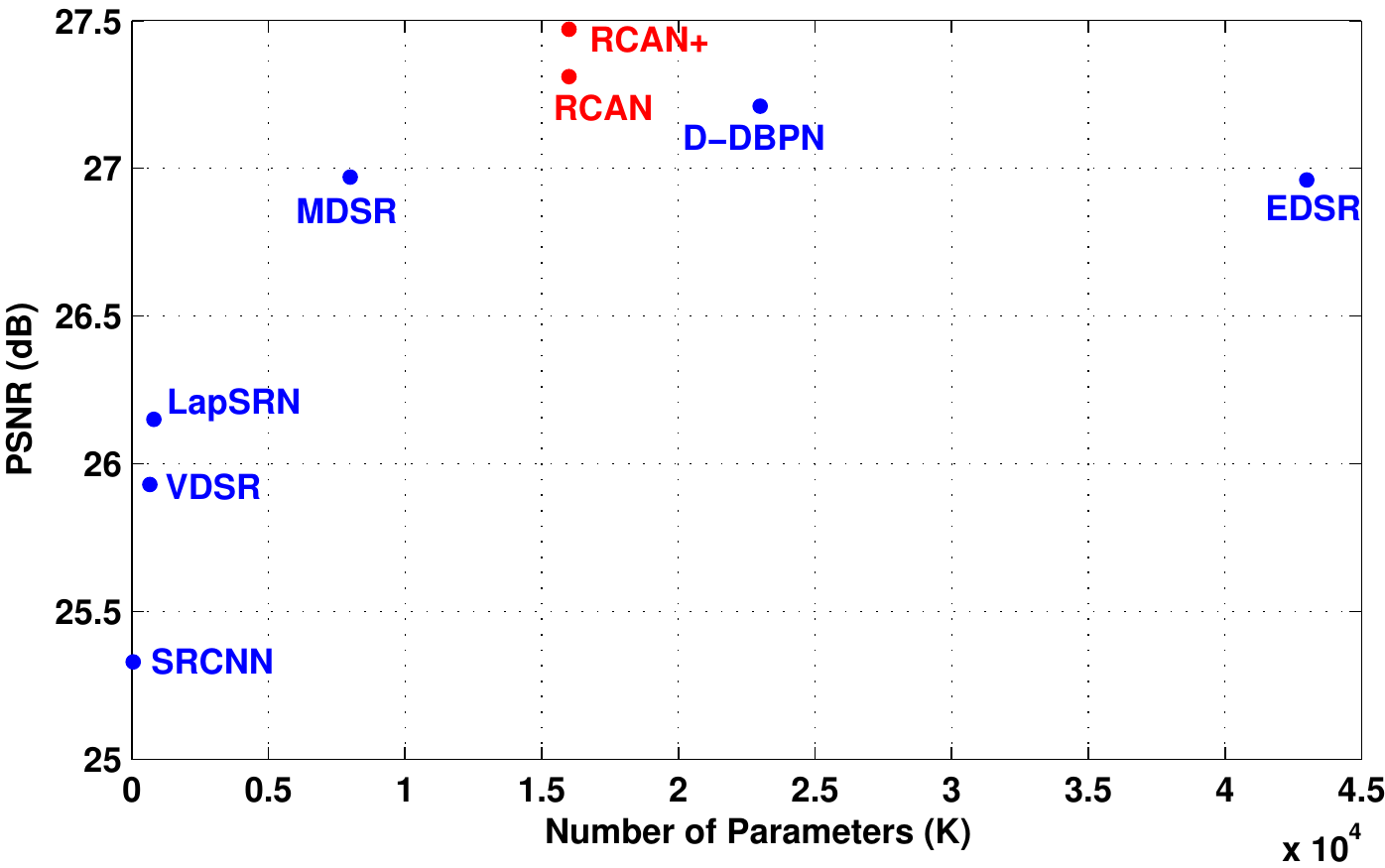}}
}
\vspace{-5mm}
\caption{Performance and number of parameters. Results are evaluated on Set5}  
\label{fig:psnr_para_x4_x8}
\vspace{-6mm}
\end{figure}

\subsection{Object Recognition Performance} 
\vspace{-2mm}
Image SR also serves as pre-processing step for high-level visual tasks (e.g., object recognition). We evaluate the object recognition performance to further demonstrate the effectiveness of our RCAN.

Here we use the same settings as ENet~\cite{sajjadi2017enhancenet}. We use ResNet-50~\cite{he2016deep} as the evaluation model and use the first 1,000 images from ImageNet CLS-LOC validation dataset for evaluation. The original cropped 224$\times$224 images are used for baseline and downscaled to 56$\times$56 for SR methods. We use 4 stat-of-the-art methods (e.g., DRCN~\cite{kim2016deeply}, FSRCNN~\cite{dong2016accelerating}, PSyCo~\cite{perez2016psyco}, and ENet-E~\cite{sajjadi2017enhancenet}) to upscale the LR images and then calculate their accuracies. As shown in Table~\ref{tab:results_object_recognition}, our RCAN achieves the lowest top-1 and top-5 errors. These comparisons further demonstrate the highly powerful representational ability of our RCAN.  
\vspace{-5mm}
\subsection{Model Size Analyses}
\vspace{-2mm}
We show comparisons about model size and performance in Figure~\ref{fig:psnr_para_x4_x8}. Although our RCAN is the deepest network, it has less parameter number than that of EDSR and RDN. Our RCAN and RCAN+ achieve higher performance, having a better tradeoff between model size and performance. It also indicates that deeper networks may be easier to achieve better performance than wider networks.

\vspace{-5mm}
\section{Conclusions}
\vspace{-3mm}
We propose very deep residual channel attention networks (RCAN) for highly accurate image SR. Specifically, the residual in residual (RIR) structure allows RCAN to reach very large depth with LSC and SSC. Meanwhile, RIR allows abundant low-frequency information to be bypassed through multiple skip connections, making the main network focus on learning high-frequency information. Furthermore, to improve ability of the network, we propose channel attention (CA) mechanism to adaptively rescale channel-wise features by considering interdependencies among channels. Extensive experiments on SR with BI and BD models demonstrate the effectiveness of our proposed RCAN. RCAN also shows promissing results for object recognition.

\noindent\textbf{Acknowledgements}: This research is supported in part by the NSF IIS award 1651902, ONR Young Investigator Award N00014-14-1-0484, and U.S. Army Research Office Award W911NF-17-1-0367.

\bibliographystyle{splncs}
\bibliography{egbib,SR_conf_bib}

\end{document}